\documentclass[]{article}
\usepackage{proceed2e}

\title{A Mixtures-of-Experts Framework for Multi-Label Classification}

\usepackage{amssymb,amsmath,dsfont}
\usepackage{bm}
\usepackage{algorithm, algorithmic}
\usepackage{graphicx}
\usepackage[round]{natbib}
\usepackage[justification=centering]{caption}

\newcommand{\tab}{\hspace*{2em}}
\newcommand{\halftab}{\hspace*{1em}}

\DeclareMathOperator*{\argmax}{arg\,max}

\author{ {\bf Charmgil Hong} \\
Computer Science Department\\
University of Pittsburgh\\
Pittsburgh, PA 15213
\And
{\bf Iyad Batal}  \\
GE Global Research\\
San Ramon, CA 94583
\And
{\bf Milos Hauskrecht}\\
Computer Science Department\\
University of Pittsburgh\\
Pittsburgh, PA 15213
}

\begin{document}\fontsize{10}{11}\rm


\maketitle

\begin{abstract}
We develop a novel probabilistic approach for multi-label classification that is based on the \textit{mixtures-of-experts} architecture combined with recently introduced \textit{conditional tree-structured Bayesian networks}.
Our approach captures different input-output relations from multi-label data using the efficient tree-structured classifiers, while the mixtures-of-experts architecture aims to compensate for the tree-structured restrictions and build a more accurate model.
We develop and present algorithms for learning the model from data and for performing multi-label predictions on future data instances. Experiments on multiple benchmark datasets demonstrate that our approach achieves highly competitive results and outperforms the existing state-of-the-art multi-label classification methods. 
\end{abstract}

\section{INTRODUCTION}
\label{sec:introduction}

Mutli-Label Classification (MLC) refers to a classification problem in which data instances are associated with multiple class variables that may reflect different views, functions or components describing the data.
MLC naturally arises in many real world problems, such as text categorization \citep{Kazawa:2005, Zhang:2006} where a document may be associated with different topics reflecting its content; semantic scene and video classification \citep{Boutell:2004:PR, Qi:2007} where different images or videos are assigned to different categories or tagged based on their content; or in genomics where individual genes may be associated with multiple functions \citep{Clare:2001:PKDD, Zhang:2006}.

Formally speaking, the MLC problem is specified by learning a function $h: \mathbb{R}^m \rightarrow \mathbf{Y}$ that maps each data instance, represented by a feature vector $\mathbf{x} = (x_1, ..., x_m)$, to class assignments, represented by a vector of $d$ binary values $\mathbf{y} = (y_1, ..., y_d)$, such that $y_i= \{ 0, 1 \}$ indicates the absence or presence of the $i$-th class. However, the application of the model in practice raises tow important questions: how to define/represent such a function for high-dimensional feature and label spaces; and how to learn this function from data.

The problem of learning MLC classifiers from data has been studied extensively by the machine learning community in recent years. 
 One of the key challenges in solving the problem is how to efficiently model and learn dependences among class variables given the fact that the number of possible class assignments is exponential in $d$. A simple solution to this is to assume that all class variables $Y_i$ for $i = 1, ..., d$ are conditionally independent of each other and, hence, learn $d$ functions to predict each class separately \citep{Clare:2001:PKDD,Boutell:2004:PR}. However, this may not suffice when dependences among labels exist.
To overcome this limitation, more advanced machine learning methods that model class relations have been proposed. 
These include two-layer classification models \citep{Godbole:2004:PAKDD,Cheng:2009:ML}, 
classifier chains \citep{Read:2009:ECML,Zhang:2010:KDD,Dembczynski:2010:ICML}, 
multi-dimensional Bayesian network classifiers \citep{Gaag:2006:PGM, Bielza:2011:IJAR,Antonucci:2013:IJCAI}
and output compression and coding methods \citep{Hsu:2009:NIPS,Tai:2010,Zhang:2011:AISTATS,Zhang:2012:ICML}.

However, the above mentioned methods are still rather limited especially when the relations among features and labels become more complex. More specifically, if the relations tend to change across the dataset, these methods may fail to respond with correct classification, since they are designed to capture only one dependence structure from data. For example, 
in automated image tagging, an object can be tagged as \{\textit{cat, pet}\} or \{\textit{cat, wild animal}\} according to its context; 
Similarly, in medical informatics, patients who are suffering from the same disease may receive different sets of medications due to their medical history or allergic reactions. To address such issues, ensemble techniques have been recently adopted to MLC settings \citep{Read:2009:ECML,Dembczynski:2010:ICML,Antonucci:2013:IJCAI}. 
However, these approaches are still limited in that they are relying on randomization for obtaining multiple dependence relations and using simple averaging to make ensemble predictions. As a result, the improvement we could obtain from them is not significant or consistent.

In this paper, we propose and study a new probabilistic MLC approach that attempts to remedy the limitations of the existing approaches. Our approach relies on the \textit{mixtures-of-experts} (ME) framework \citep{Jacobs:1991:AML,Yuksel:2012:IEEE} and \textit{conditional tree-structured Bayesian network} (CTBN) classifiers proposed recently in \citep{batal:2013:CIKM}.
Briefly, CTBN defines a multi-label classifier that is modeling $P(\mathbf{Y}|\mathbf{X})$ where dependences among class variables for different inputs are modeled by a collection of classifiers (for example, logistic regression models) linked together in directed tree structures. 
The model comes with a number of computational advantages such as efficient learning, and efficient MAP inference that lets us find the best set of class assignments for a given data instance $\mathbf{x}$. 

A limitation of CTBN is that dependences among class variables are restricted to tree structures which may not reflect all existing dependences in the data. Our new framework based on the ME architecture aims to take advantage of the computational benefits of CTBNs and remedy its limitation by learning and combining multiple CTBNs, where each CTBN can cover a different region of the input space and/or can help to model the different dependences among class variables. We develop and present an EM algorithm for learning the new model from data and an algorithm for making the MAP inferences for predicting class assignments for future data instances.

The rest of the paper is organized as follows.
Section \ref{sec:problem_definition} formally defines the MLC problem.
Section \ref{sec:related_research} gives summary on related MLC research.
Section \ref{sec:preliminary} provides the necessary definitions for ME and CTBN, which are needed to understand our model.
Section \ref{sec:proposed} describes our proposed solution by going over its model representation (Section \ref{subsec:representation}) and the supporting algorithms for parameter learning (Section \ref{subsec:param_learn}), structure learning (Section \ref{subsec:struct_learn}) and prediction (Section \ref{subsec:predict}).
Section \ref{sec:experiments} presents the experimental results and evaluation.
Section \ref{sec:conclusion} concludes the paper.

\section{PROBLEM DEFINITION}
\label{sec:problem_definition}

Multi-Label Classification (MLC) is a classification problem in which each data instance is associated with a subset of labels from a set of possible labels $L$. Let $d = |L|$. We can define $d$ binary class variables $Y_1, ..., Y_d$, where the value of $Y_i$ in instance $\mathbf{x}$ indicates whether or not the $i$-th label in $L$ is present in $\mathbf{x}$. 
We are given labeled training data $D=\{\mathbf{x}^{(n)}, \mathbf{y}^{(n)}\}_{n=1}^N$, where $\mathbf{x}^{(n)}=(x^{(n)}_1,...,$ $x^{(n)}_m)$ is the \emph{m}-dimensional feature vector of the $n$-th instance (the input) and $\mathbf{y}^{(n)}=(y^{(n)}_1, ..., y^{(n)}_d)$ is its \emph{d}-dimensional class vector (the output). 
We want to learn a function $h$ that fits $D$ and assigns to each instance, represented by its feature vector, a class vector:
$$h: \mathbb{R}^m \rightarrow \{ 0, 1 \}^d$$
One way to approach this task is to model and learn the \textit{conditional joint distribution} $P(\mathbf{Y|X})$, where $\mathbf{Y}$ is a random variable for the class vector and $\mathbf{X}$ is a random variable for the feature vector. Assuming the 0-1 loss function, the optimal classifier $h^*$ assigns to each instance $\mathbf{x}$ the maximum a posteriori (MAP) assignment of class variables:
\begin{align}
\label{OPT-classification}
h^*(\mathbf{x}) &= \argmax_{\mathbf{y}} P(\mathbf{Y\!=\!y|X\!=\!x})\\
                &= \argmax_{y_1, ..., y_d} P(Y_1\!=\!y_1, ..., Y_d\!=\!y_d|\mathbf{X\!=\!x})\notag
\end{align}
A key challenge for modeling and learning $P(\mathbf{Y|X})$ from data, as well as for defining the corresponding MAP classifier, is that the number of all possible class assignments for a given $\mathbf{x}$ is exponential in $d$ (there are $2^d$ different assignments). Our goal is to develop a parametric model that allows us to efficiently model and learn $P(\mathbf{Y|X})$ from data.

\noindent
\textbf{Notation:} For notational convenience, we will omit the index superscript $^{(n)}$ when it is not necessary. We may also abbreviate the expressions by omitting variable names; e.g., $P(Y_1\!=\!y_1, ..., Y_d\!=\!y_d|\mathbf{X\!=\!x}) = P(y_1, ..., y_d|\mathbf{x})$.

\section{RELATED RESEARCH}
\label{sec:related_research}

In this section, we review the related research in MLC and outline the main differences from our approach.

Earlier MLC methods ignore the relations between classes and learn to predict each class separately \citep{Clare:2001:PKDD,Boutell:2004:PR}. \citet{Zhang:2007:PR} presented the multi-label k-nearest neighbor method, which predicts each class label by combining KNN with Bayesian inference. An approach that enriches the feature space by incorporating an intermediate layer of classifiers was proposed by \citep{Godbole:2004:PAKDD} and later on by \citep{Cheng:2009:ML}. The main drawback of these methods is that class dependences are either not represented at all, or represented indirectly in a limited way.

Several methods have been proposed to probabilistically model MLC.
The classifier chains (CC) method \citep{Read:2009:ECML} decomposes the relations among the class variables using the chain rule of probability:
\begin{align*}
P(Y_1, ..., Y_d | \mathbf{X}) = \prod_{i=1}^d P(Y_i | \mathbf{X}, Y_1, ..., Y_{i-1})
\end{align*}
Each component in the chain is a classifier that is learned separately by incorporating the (0/1) predictions of preceding classifiers as additional features.
\citet{Zhang:2010:KDD} further studied the influence of the chain order and presented a method to learn an effective ordering from data.
The main disadvantage of CC, however, is that they do not perform proper probabilistic inference for classification. Instead, they simply propagate the predictions according to the class order defined by the chain.
To overcome this shortcoming, \citet{Dembczynski:2010:ICML} presented the probabilistic classifier chains method by extending CC to estimate the entire class posterior distribution. However, this method needs to evaluate all possible $2^d$ label configurations, which greatly limits its applicability.

Other probabilistic MLC methods are based on multi-dimensional Bayesian networks \citep{Gaag:2006:PGM,Bielza:2011:IJAR,Antonucci:2013:IJCAI}. These methods attempt to build a generative model of $P(\mathbf{X,Y})$ using a restricted Bayesian network structure, which assumes all class variables are top nodes and all feature variables are their descendants. The limitation of this approach is that it must learn the structure of the full joint distribution over both features and classes, which can be very complex. Besides, it requires all features to be a priori discretized. In contrast, our approach directly learns the conditional distribution $P(\mathbf{Y|X})$ and takes advantage of modern discriminative classifiers.


An alternative approach for MLC is based on output coding. 
The idea is to project the output space $\mathbf{Y}$ into a lower dimensional space $\mathbf{Y'}$, learn to predict $\mathbf{Y'}$, and then reconstruct the original output from the noisy predictions.
Methods that fall into this category use different dimensionality reduction techniques, such as compressed sensing \citep{Hsu:2009:NIPS}, principal component analysis \citep{Tai:2010} and canonical correlation analysis \citep{Zhang:2011:AISTATS}. The state-of-the-art in output coding utilizes a maximum margin formulation \citep{Zhang:2012:ICML} that promotes both discriminative and predictable codes.
The limitation of output coding methods is that they can only predict the single ``best'' output for a given input, and they cannot compute probabilities for different input-output pairs.

Several researchers proposed using ensemble methods for MLC. In general, the objective was to compensate for the restrictions the base MLC models introduce using a set of models and their combinations. \cite{Read:2009:ECML} presented a simple method that averages the predictions of multiple randomly ordered CCs trained on random subsets of the data.
\cite{Antonucci:2013:IJCAI} proposed an ensemble of multi-dimensional Bayesian networks combined via simple averaging. 
In this work, we develop an approach based on the mixtures-of-experts architecture. The difference from the previous work is that our approach optimizes the structures and parameters of base classifiers in a principled way, and that our approach can help to overcome the restriction of the base MLC classifier by modeling better the relations among inputs and their labels, as well as, mutual relations among different labels.

\section{PRELIMINARY}
\label{sec:preliminary}

The MLC solution we propose in this work combines multiple base MLC classifiers using the \textit{mixtures-of-experts} (ME) \citep{Jacobs:1991:AML} architecture. The base classifiers we use are based on the \textit{conditional tree-structured Bayesian networks} (CTBN) \citep{batal:2013:CIKM}. To start with, we briefly review the basics of ME and CTBN.

ME is a mixture model that consists of a set of \textit{experts} that are combined via \textit{gating (or switching)} module to represent the conditional distribution $P(y|\mathbf{x})$. The model is defined by the following decomposition:
\begin{align}
P(y | \mathbf{x}) 
	\label{eq:def-me}
	&= \sum_{k=1}^K P(E_k|\mathbf{x}) P(y | \mathbf{x}, E_k),
\end{align}
where $P(y | \mathbf{x}, E_k)$ is the distribution of outputs defined by the $k$-th expert $E_k$
and $P(E_k|\mathbf{x})$ is the context sensitive prior of the $k$-th expert that is implemented by the gating function $g_k(\mathbf{x})$. In general, depending on the choice of the expert model, ME can be used for either regression or classification \citep{Yuksel:2012:IEEE}. 

The ME model defines a soft-partitioning of the input space (via the gating module and its functions), on which the $K$ experts represent different input-output relations. ME is especially useful when individual expert models are good in representing local input-output relations but fail to accurately capture the relations for the complete input space. The ability to switch among the experts in different regions of the input space allows to compensate for the limitation of individual experts and improve the overall model and its accuracy. 
In general, depending on the choice of the expert model, ME can be used for either regression or classification \citep{Yuksel:2012:IEEE}. 


ME has been successfully adopted in a range of applications 
such as handwriting recognition \citep{Ebrahimpour:2009:JDCTA}, text classification \citep{Estabrooks:2001}, 
climate prediction \citep{Lu:2006:PR} and bioinformatics \citep{Qi:2007:BMC,Cao:2010}.
In addition, ME has been vigorously used in time series analysis, including speech recognition \citep{Mossavat:2010}, financial forecasting \citep{Weigend:2000} and dynamic control systems \citep{Jacobs:1993,Weigend:1995}. 
Recently, ME has been used in social network analysis, in which different social behavior patterns are modeled through mixtures \citep{Gormley:2011}. 

In this work, we apply the ME approach in context of MLC. We would like to note that although modeling of the joint conditional probability $P(y_1, ..., y_d | \mathbf{x})$ has been attempted in context of multiple target regression \citep{Jordan:1995,Waterhouse:thesis:1997}, its application to MLC is new to the best of our knowledge. 

\begin{figure}[t]
\centering
\includegraphics[width=.3\textwidth]{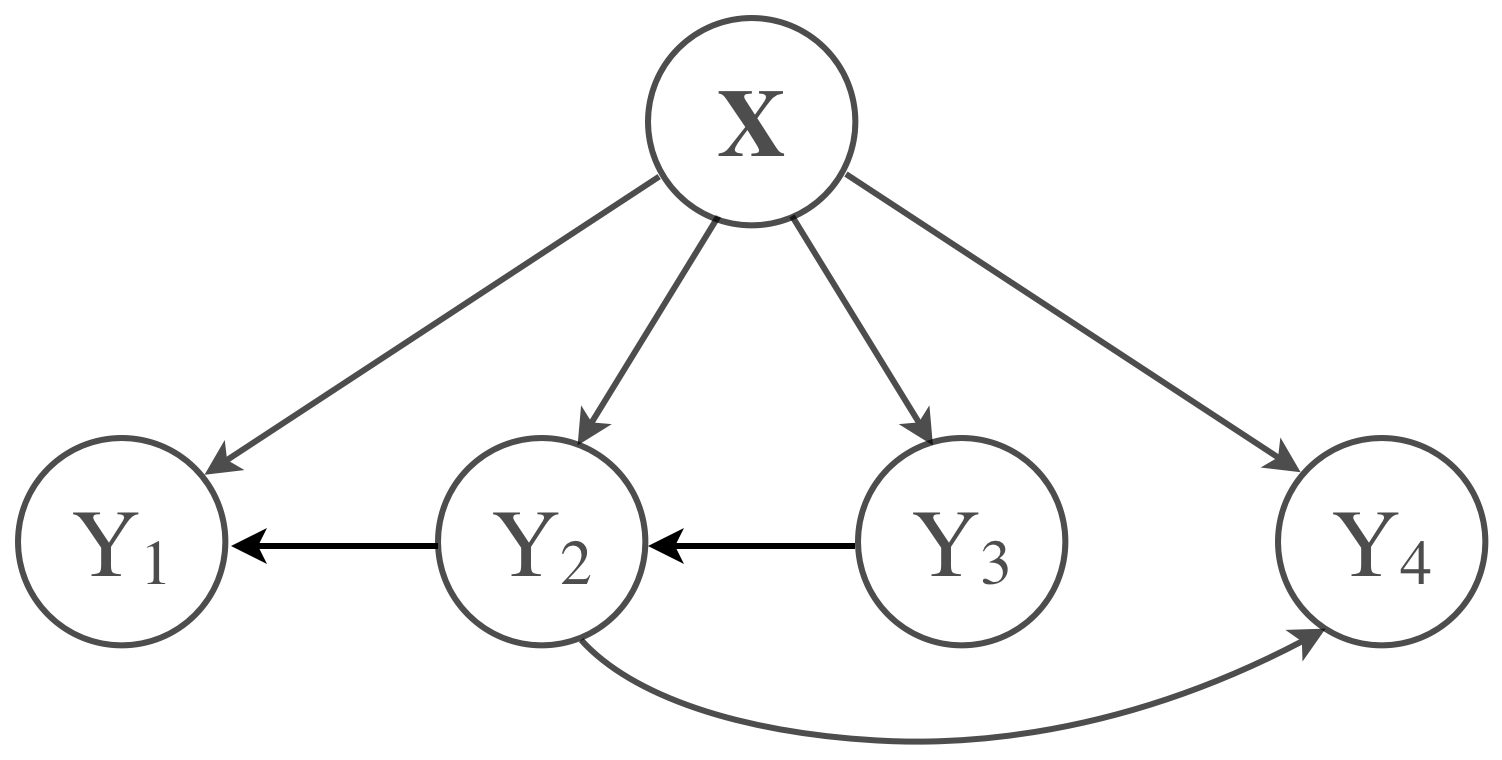}
\caption{\footnotesize An example of CTBN.}
\label{fig:CTBN-example}
\end{figure}

In particular, we combine ME with CTBN to model individual experts.
CTBN is a recently proposed probabilistic MLC method that has been shown to be competitive and efficient on a range of domains.
CTBN defines $P(\mathbf{Y} | \mathbf{X})$ using a collection of classifiers modeling relations in between features and individual labels that are tied together using a special Bayesian network structure that approximates the dependence relations among the class variables. In modeling of the dependences, it allows each class variable to have at most one other class variable as a parent (without creating a cycle) besides the feature vector $\mathbf{X}$. 

A CTBN $T$ defines the joint distribution of class vector $(y_1,...,y_d)$ conditioned on feature vector $\mathbf{x}$ as:
\begin{align}
\label{eq:def-ctbn}
P(y_1,...,y_d | \mathbf{x}, T)
	& = \prod^d_{i=1} P(y_i | \mathbf{x}, y_{\pi(i, T)}),
\end{align}
where $\pi(i, T)$ denotes the parent class of class $Y_i$ in $T$ (by convention, $\pi(i,T)=\{\}$ if $Y_i$ does not have a parent class). For example, the conditional joint distribution of class assignment $(y_1, y_2, y_3, y_4)$ given $\mathbf{x}$ according to the network $T$ in Figure \ref{fig:CTBN-example} is defined as:
\begin{align*}
&P(y_1, y_2, y_3, y_4|\mathbf{x}, T) \\
&\tab= P(y_3 | \mathbf{x}) \cdot P(y_2 | \mathbf{x}, y_3) \cdot P(y_1 | \mathbf{x}, y_2) \cdot P(y_4 | \mathbf{x}, y_2)
\end{align*}

\section{PROPOSED SOLUTION}
\label{sec:proposed}

In this section, we develop a \textit{Multi-Label Mixtures-of-Experts} (ML-ME) model, which uses the ME framework in combination with the CTBN classifiers to improve the classification accuracy of MLC tasks, and develop algorithms for its learning and predictions. 
Our key motivation is to exploit the divide and conquer principle, which states that a large, complex problem can be decomposed and effectively solved using simpler sub-problems. More specifically, we want to accurately model relations among inputs $\mathbf{X}$ and class variables $\mathbf{Y}$ by learning multiple CTBN models and by improving their predictive ability by combining their outputs.
In section \ref{subsec:representation}, we describe the mixture defined by the ML-ME model. In section \ref{subsec:param_learn} through \ref{subsec:predict}, we present the learning and prediction algorithms for the ML-ME model.

\subsection{REPRESENTATION}
\label{subsec:representation}

By following the definition of ME in Equation (\ref{eq:def-me}), ML-ME defines the multivariate posterior distribution of class vector $\mathbf{y} = (y_1, ..., y_d)$ as:
\begin{align}
P(\mathbf{y} | \mathbf{x}) 
	\label{eq:def-ml-me}
	&= \sum_{k=1}^K g_k(\mathbf{x}) P(\mathbf{y} | \mathbf{x}, E_k),
\end{align}
where $P(\mathbf{y} | \mathbf{x}, E_k)$ is the joint conditional distribution defined by the $k$-th expert $E_k$
and $g_k(\mathbf{x})\!=\!P(E_k|\mathbf{x})$ is the gating function
 reflecting how much the $k$-th expert should contribute to predict classes for input $\mathbf{x}$.
In this work, we model the gating functions for the different experts with the help of the softmax function, which is also know as normalized exponential:
\begin{align}
g_k(\mathbf{x})
	\label{eq:def-gate}
	&= \frac{\exp(\bm\theta_{G_k}\mathbf{x})}{\sum_{k'=1}^K \exp(\bm\theta_{G_{k'}}\mathbf{x})},
\end{align}
where $\bm\Theta_G = \{ \bm\theta_{G_k} \}_{k=1}^K$ is a set of the softmax parameters.

\begin{figure}[t]
\centering
\includegraphics[width=.45\textwidth]{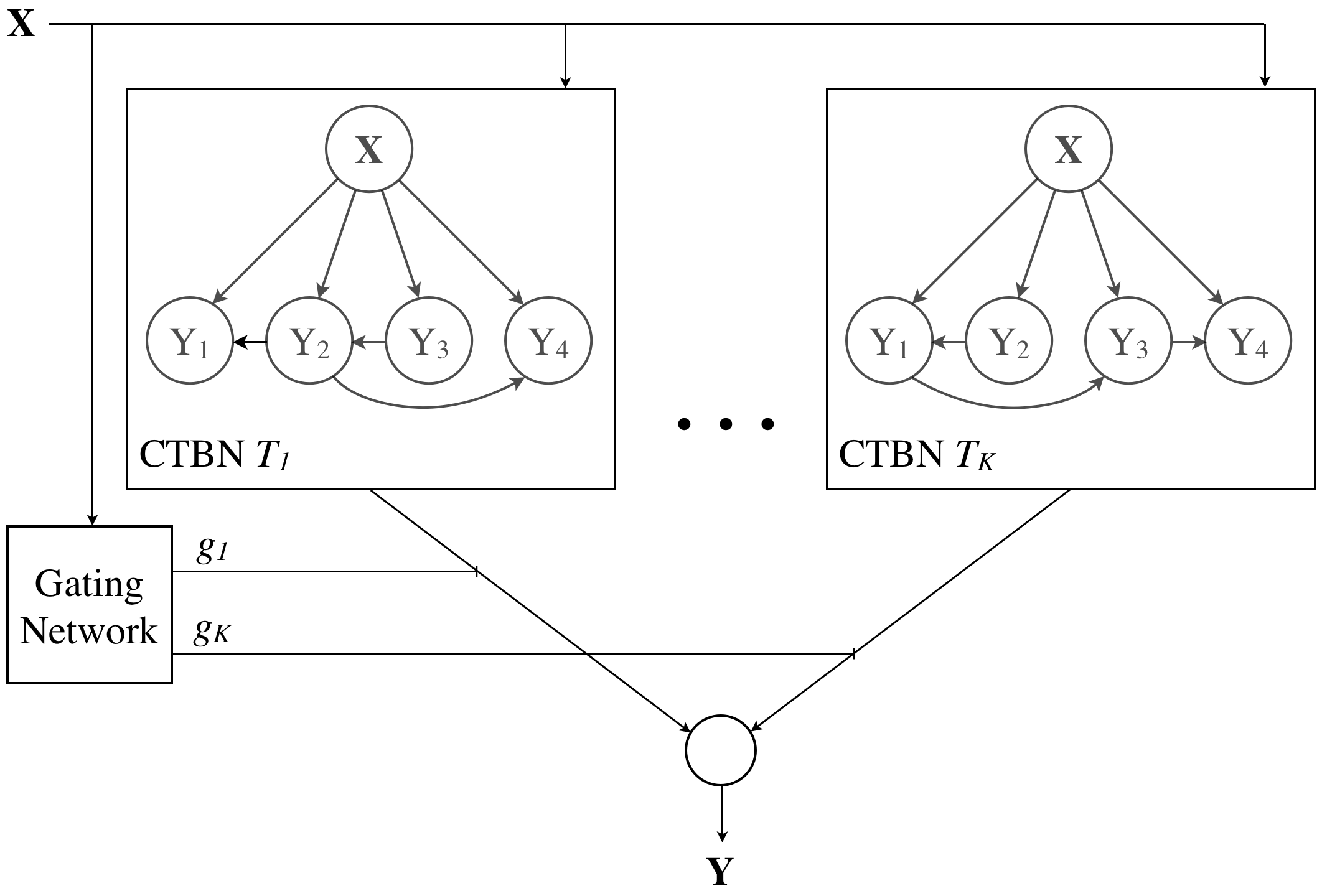}
\caption{\footnotesize An example of ML-ME.}
\label{fig:MLME-example}
\end{figure}

To model the different experts and their probabilistic MLC predictions $P(\mathbf{y} | \mathbf{x}, E_k)$ in Equation (\ref{eq:def-ml-me}), we rely on the CTBN model introduced in the previous section.
Employing multiple CTBN models $T_k : k \in \{1, ..., K\}$ as experts in the mixture, our ML-ME model defines the joint distribution of class vector $\mathbf{y}$ as:
\begin{align}
P(\mathbf{y} | \mathbf{x}) 
	\label{eq:def-ml-me-ctbn}
	&= \sum_{k=1}^K g_k(\mathbf{x}) P(\mathbf{y} | \mathbf{x}, T_k)\\
	&= \sum_{k=1}^K g_k(\mathbf{x}) \prod^d_{i=1} P(y_i | \mathbf{x}, y_{\pi(i, T_k)}),\notag
\end{align}
Figure \ref{fig:MLME-example} depicts an example ML-ME model, which consists of $K$ CTBNs and whose output is probabilistically mixed with the help of the gating network.

\textbf{Parameters}\halftab 
Let $\bm\Theta = \{ \bm\Theta_G, \bm\Theta_T \}$ denote the set of all the parameters of the ML-ME model, where $\bm\Theta_{G} = \{ \bm\theta_{G_k} \}_{k=1}^K$ are parameters of the the gating model, and $\bm\Theta_{T} = \{ \bm\theta_{T_k} \}_{k=1}^K$  are parameters of CTBNs defining individual experts. Since the gating function for each expert is defined by a linear combination of inputs, there are $|\bm\theta_{G_k}| = (m+1)$ parameters per expert. Consequently,
the total number of parameters needed to represent the full gating network is: $|\bm\Theta_{G}| = K(m+1)$.  

The parameterization of a CTBN expert is done by modeling the conditional probability distributions (CPDs) of class variable $Y_i$ conditioned on its parents. For each $Y_i$, we learn two logistic regression classifier functions according to the parent class: $P(y_i | \mathbf{x}, y_{\pi(i, T_k)}\!=\!0)$ and $P(y_i | \mathbf{x}, y_{\pi(i, T_k)}\!=\!1)$. Let $\bm\theta_{T_k}$ denote the parameters for the $k$-th CTBN expert. 
Then, for $|\bm\theta_{T_k}| = 2d(m+1)$, since two different classifiers are defined on each class variable. Hence, $|\bm\Theta_{T}| = 2dK(m+1)$.

Table \ref{table:notations} summarizes the notation and parameterization of our ML-ME model. In summary, the total number of parameters for our model is $K(1+2d)(m+1)$.

\begin{table}[t]
\caption{Notations}
\label{table:notations}
\footnotesize
\begin{center}
	\begin{tabular}{cl}
		\multicolumn{1}{c}{\bf NOTATION}  &\multicolumn{1}{c}{\bf DESCRIPTION} \\
		\hline \\
		$m$ & Input (feature) dimensionality\\
		$d$ & Output (class) dimensionality\\
		$N$ & Number of data instances\\
		$K$ & Number of experts in a mixture\\
		$T_k$ & CTBN with index $k$\\
		$\bm\Theta_T = \{ \bm\theta_{T_1}, ..., \bm\theta_{T_K}\}$ & The parameters for CTBNs\\
		$\bm\Theta_G = \{ \bm\theta_{G_1}, ..., \bm\theta_{G_K}\}$ & The parameters for a gate\\\\
		\hline
	\end{tabular}
\end{center}
\end{table}

\subsection{PARAMETER LEARNING FOR FIXED STRUCTURES}
\label{subsec:param_learn}

In this section, we describe how to learn the parameters of ML-ME, when the structures of individual CTBN experts are known and fixed. We return to the structure learning problem in Section \ref{subsec:struct_learn}.

$\bm\Theta = \{ \bm\Theta_G, \bm\Theta_T \}$ denotes the set of all the parameters of the ML-ME model. Our objective is to find the parameters that optimize the log-likelihood of the training data:
$$l(D; \bm\Theta) = \sum_{n=1}^N\log P(\mathbf{y}^{(n)}|\mathbf{x}^{(n)})$$
By substituting the joint probability with the definition of ML-ME (Equation (\ref{eq:def-ml-me-ctbn})), we obtain:
\begin{align}
\label{eq:oll}
l(D; \bm\Theta)
	&= \sum_{n=1}^N \log\sum_{k=1}^K g_k(\mathbf{x}^{(n)}) P(\mathbf{y}^{(n)}|\mathbf{x}^{(n)},T_k)
\end{align}
We refer to Equation (\ref{eq:oll}) as the \textit{observed log-likelihood}. However, optimizing this function is very difficult because there is a summation inside the $\log$, which results in a non-convex function. To overcome this difficulty, we instead optimize the \textit{complete log-likelihood} in the expectation-maximization (EM) framework.

The complete log-likelihood is defined by associating each instance $(\mathbf{x}^{(n)}, \mathbf{y}^{(n)})$ with a hidden variable $z^{(n)} \in \{1,...,K\}$ indicating to which expert it belongs to:
\begin{align}
\label{eq:param_learn1}
&l_c(D; \bm\Theta)
	=\sum_{n=1}^N \log P(\mathbf{y}^{(n)}, z^{(n)} | \mathbf{x}^{(n)}) \\
	&= \sum_{n=1}^N \log \prod_{k=1}^K P(\mathbf{y}^{(n)}, T_k | \mathbf{x}^{(n)})^{\mathds{1}[z^{(n)} = k]} \notag\\
	&= \sum_{n=1}^N \log \prod_{k=1}^K \left[ g_k(\mathbf{x}^{(n)}) P(\mathbf{y}^{(n)} | \mathbf{x}^{(n)}, T_k) \right]^{\mathds{1}[z^{(n)} = k]} \notag\\
	&= \sum_{n=1}^N \sum_{k=1}^K {\mathds{1}[z^{(n)} = k]} \left[ \log g_k(\mathbf{x}^{(n)}) P(\mathbf{y}^{(n)} | \mathbf{x}^{(n)}, T_k) \right]\notag,
\end{align}
where $\mathds{1}[z^{(n)} = k]$ is the indicator function that evaluates to one if the $n$-th instance belongs to the $k$-th expert and to zero otherwise.

The EM framework iteratively optimizes the \textit{expected complete log-likelihood} $(E\left[l_c(D; \bm\Theta)\right])$, which is always a lower bound of the observed log-likelihood \citep{Dempster:1977}. In the \textit{E-step}, the expectation is computed using the current set of parameters; in the \textit{M-step}, the parameters of the model are relearned by maximizing the expected complete log-likelihood. In the following, we explain our parameter learning algorithm by deriving the E-step and M-step for ML-ME.

\textbf{E-step}\halftab In the E-step, we compute the expectation of the complete log-likelihood, which reduces to computing the expectation of the hidden variable.
\begin{align*}
E\left[{\mathds{1}[z^{(n)}\!= k]}\right] 
	& = P(z^{(n)}\!= k | \mathbf{y}^{(n)}, \mathbf{x}^{(n)})\\
	& = P(T_k|\mathbf{y}^{(n)}, \mathbf{x}^{(n)})
\end{align*}
Notice that this becomes the posterior of the $k$-th expert given the observation and the current set of parameters.
Let $h_k^{(n)}$ denote $P(T_k|\mathbf{y}^{(n)}, \mathbf{x}^{(n)})$. We can write $h_k^{(n)}$ using Bayes rule as:
\begin{align}
h_k^{(n)}
	\label{eq:estep_3}
	&= \frac{g_k(\mathbf{x}^{(n)}) P(\mathbf{y}^{(n)}|\mathbf{x}^{(n)},T_k)}{\sum_{k'=1}^Kg_{k'}(\mathbf{x}^{(n)}) P(\mathbf{y}^{(n)}|\mathbf{x}^{(n)},T_{k'})}
\end{align}

\textbf{M-step}\halftab In the M-step, we learn the model parameters $\{ \bm\Theta_G, \bm\Theta_T \}$ that maximize the expected complete log-likelihood. Let us first rewrite Equation (\ref{eq:param_learn1}) using $h_k^{(n)}$ and by switching the order of summations:

\vspace{-0.15in}
\begin{small}
\begin{align}
\label{eq:mstep_1}
&\sum_{k=1}^K \sum_{n=1}^N h_k^{(n)} \log g_k(\mathbf{x}^{(n)}) + h_k^{(n)} \log P\left(\mathbf{y}^{(n)} | \mathbf{x}^{(n)}, T_k\right)
\end{align}
\end{small}
For the fixed $h_k^{(n)}$ we can decompose Equation (\ref{eq:mstep_1}) into two parts, each of which respectively involves the gating model parameters $\bm\Theta_G$ and the CTBN parameters $\bm\Theta_T$:
\begin{align*}
&f_1(D;\bm\Theta_G) = \sum_{k=1}^K \sum_{n=1}^N h_k^{(n)} \log g_k(\mathbf{x}^{(n)}) \\
&f_2(D;\bm\Theta_T) = \sum_{k=1}^K \sum_{n=1}^N h_k^{(n)} \log P\left(\mathbf{y}^{(n)} | \mathbf{x}^{(n)}, T_k\right)
\end{align*}
Notice that, in order to optimize the gate parameters $\bm\Theta_G$, we only need to optimize $f_1(D;\Theta_G)$; whereas to optimize the CTBN parameters $\bm\Theta_T$, we only need to optimize $f_2(D;\bm\Theta_T)$. 

We first show how to learn the parameters $\bm\Theta_G = \{ \bm\theta_{G_1}, ..., \bm\theta_{G_K} \}$ of the gating model. We can rewrite $f_1(D;\bm\Theta_G)$ using Equation (\ref{eq:def-gate}) as:

\vspace{-0.15in}
\begin{small}
\begin{align*}
&f_1(D;\bm\Theta_G)\\
&= \sum_{k=1}^K\sum_{n=1}^N h_k^{(n)} \bm\theta_{G_k}\mathbf{x}^{(n)} - h_k^{(n)} \log \sum_{k'=1}^K\exp(\bm\theta_{G_{k'}}\mathbf{x}^{(i)}) 
\end{align*}
\end{small}

\vspace{-0.15in}
Since $f_1(D;\bm\Theta_G)$ is concave in $\bm\Theta_G$, gradient methods are guaranteed to find its optimal solution. The derivative of the 
log-likelihood with respect to $\bm\theta_{G_j}$ is calculated as:
\begin{align}
\nabla_{\theta_j} f_1(D;\bm\Theta_G)
	\label{eq:derive_mstep1}
	&= \sum_{n=1}^N \left\{  h_j^{(n)} - g_j(\mathbf{x}^{(n)})  \right\} \mathbf{x}^{(n)} 
\end{align}
Note that the derivative becomes zero when $g_j(\mathbf{x}^{(n)}) = P(T_k|\mathbf{x}^{(n)})$ and $h_j^{(n)} = P(T_k|\mathbf{y}^{(n)}, \mathbf{x}^{(n)})$ are equal.

We can solve the optimization of $f_1(D;\bm\Theta_G)$ using any gradient method. However, in practice the dimensionality of the input space can be high which may result in model overfitting. To prevent the overfitting problem, we add $L_2$-regularization penalty $R(\bm\Theta_G) = \frac{\lambda}{2} \sum_{k=1}^K ||\bm\theta_{G_k}||_2^2$ to the optimization function to penalize model complexity. In our experiments, we use the L-BFGS algorithm \citep{Liu:1989} to optimize this regularized objective function. L-BFGS is a quasi-Newton optimization method that uses a sparse approximation to the inverse Hessian matrix to steer its search through parameter space. The algorithm is known to provide faster convergence rate and be well suited for optimization problems with a large number of variables.

To optimize the parameters of the CTBN experts $\bm\Theta_T = \{ \bm\theta_{T_1}, ..., \bm\theta_{T_K} \}$, we optimize $f_2(D;\bm\Theta_T)$.  This optimization decomposes into optimization of parameters of individual CTBN $T_k$. Note that $f_2$ is the weighted log-likelihood where $h_k^{(n)}$ serves as the instance weight. 
Following \citep{batal:2013:CIKM}, we use the logistic regression model to represent $P(\mathbf{y}^{(n)}|\mathbf{x}^{(n)}, T_k)$ in all our experiments. To prevent overfitting, we apply 
$L_2$-regularized instance-weighted logistic regression to optimize $f_2$. Algorithm \ref{alg:learn-ML-ME-parameters} summarizes our parameter learning algorithm.

\begin{algorithm}[tb]
    \caption{\footnotesize learn-mixture-parameters}
    \label{alg:learn-ML-ME-parameters}
    \mbox{\footnotesize \textbf{Input}: Training data $D$; base CTBN experts $T_1, ..., T_K$}
    \mbox{\footnotesize \textbf{Output}: Model parameters $\{ \bm\Theta_G, \bm\Theta_T \}$}
\vspace{-0.15in}
{\footnotesize
\begin{algorithmic}[1]
    \REPEAT
        \STATE {\bfseries E-step:}
        \FOR{$k=1$ {\bfseries to} $K$, $n=1$ {\bfseries to} $N$}
            \STATE Compute $h_k^{(n)}$ using Equation (\ref{eq:estep_3})
        \ENDFOR
        \STATE {\bfseries M-step:}
        \STATE $\bm\Theta_{G} = \argmax_{\bm\Theta_{G}} l_1(D;\bm\Theta_G) - R(\bm\Theta_G)$
        \FOR{$k=1$ {\bfseries to} $K$}
            \STATE $\bm\theta_{T_k} = \displaystyle\argmax \textstyle\sum_{n=1}^N h_k^{(n)} \log P(\mathbf{y}^{(n)} | \mathbf{x}^{(n)}, T_k)$
        \ENDFOR
    \UNTIL{\emph{convergence}}
\end{algorithmic}
}
\end{algorithm}

\subsubsection{Complexity}
\textbf{E-step}\halftab 
We compute $h_k^{(n)}$ for each instance on every CTBN expert. This requires $O(md)$ multiplications. Hence, the complexity of a single E-step is $O(KNmd)$.

\textbf{M-step}\halftab 
Computing the derivative in Equation (\ref{eq:derive_mstep1}) requires $O(mN)$ multiplications, therefore optimizing $\bm\Theta_G$ requires $O(mNl)$ operations, where $l$ is the number of L-BFGS steps. To learn $\bm\Theta_T$, we optimize an instance-weighted logistic regression for every node of  each of the $K$ experts, which requires learning of $O(Kd)$ logistic regression models.


\subsection{STRUCTURE LEARNING}
\label{subsec:struct_learn}

In the previous section, we described the parameter learning of ML-ME by assuming we have fixed the individual CTBN structures. In this section, we present how to automatically learn CTBN structures from data. In a nutshell, we apply a sequential boosting-like heuristic; i.e., on each iteration, we learn a structure that focuses on ``hard instances" that previous CTBNs tend to misclassify. In the following, we first describe how to learn a single CTBN structure from instance-weighted data. After that, we describe how to re-weight the instances and incrementally add new structures to the ML-ME model.

\subsubsection{Learning a Single CTBN Structure on Weighted Data}
\label{subsubsec:learning-single-structure}

To learn the CTBN structure that best approximates weighted data, we find the structure that maximizes the weighted conditional log-likelihood (WCLL) on $\{D, \Omega\}$, where $D= \{\mathbf{x}^{(n)},\mathbf{y}^{(n)}\}_{n=1}^N$ is the data and $\Omega= \{\omega^{(n)}\}_{n=1}^N$ is the instance weight. Note that we further split $D$ into training data $D_{tr}$ and hold-out data $D_h$. 

Given a CTBN structure $T$, we train its parameters using $D_{tr}$, which corresponds to learning instance-weighted logistic regression using $D_{tr}$ and the corresponding instance weights.
On the other hand, we use WCLL of $D_h$ to define the score that measures the quality of $T$.

\vspace{-0.15in}
\begin{small}
\begin{align}
	Score(T)
		&=\sum_{n \in D_h} \omega^{(n)} \log P(\mathbf{y}^{(n)}|\mathbf{x}^{(n)},T)\notag
		\label{eq:structure_score_1} \\
		&= \sum_{n \in D_h} \sum_{i=1}^d \omega^{(n)} \log P(y^{(n)}_i | \mathbf{x}^{(n)}, y^{(n)}_{\pi(i,T)})
\end{align}
\end{small}
\vspace{-0.15in}

Below we describe our algorithm for obtaining the CTBN structure that optimizes Equation (\ref{eq:structure_score_1}) without having to evaluate all of the exponentially many possible tree structures.


Let us first define a weighted directed graph $G = (V,E)$ as follows:

\vspace{-0.175in}
\begin{itemize}
	\itemsep-.1em
	\item There is one vertex $V_i$ for each class variable $Y_i : i \in \{1,...,d\}$.
	\item There is a directed edge $E_{j \rightarrow i}$ from each vertex $V_j$ to each vertex $V_i$ (i.e., $G$ is complete). In addition, each vertex $V_i$ has a self-loop $E_{i \rightarrow i}$.
	\item The weight of edge $E_{j \rightarrow i}$, denoted as $W_{j \rightarrow i}$, is the WCLL of class $Y_i$ conditioned on $\mathbf{X}$ and $Y_j$:
	
		\vspace{-0.2in}
		\begin{small}
		\begin{align}
		\label{eq:structure_score_2}
		W_{j \rightarrow i} = \sum_{n \in D_h} \omega^{(n)} \log P(y_i^{(n)}|\mathbf{x}^{(n)}, y_j^{(n)})
		\end{align}
		\end{small}
	
	\vspace{-0.2in}
	\item The weight of self-loop $E_{i \rightarrow i}$, denoted as $W_{\phi \rightarrow i}$, is the WCLL of class $Y_i$ conditioned only on $\mathbf{X}$.
\end{itemize}
Using the definition of edge weights (Equation (\ref{eq:structure_score_2})) and by switching the order of the summations in Equation (\ref{eq:structure_score_1}), we can rewrite the score of $T$ simply as the sum of its edge weights:
%
\begin{align*}
	Score(T) &= \sum_{n=1}^d W_{\pi(i,T) \rightarrow i}
\end{align*}
%
Now we have transformed the problem of finding the optimal tree structure into the problem of finding the tree in $G$ that has the maximum sum of edge weights. The solution can be obtained by solving the maximum branching (arborescence) problem \citep{Edmonds:1967}, which finds the maximum weight tree in a weighted directed graph.

\subsubsection{Learning Multiple CTBN Structures}
\label{subsubsec:learning-multiple-structures}

In order to obtain multiple, effective CTBN structures for the ML-ME model, we apply the above described algorithm multiple times with different sets of instance weights. We assign the weights such that we give poorly predicted instances higher weights; and give well-predicted instances lower weights.

We start with assigning all instances uniform weights (i.e., all instances are equally important a priori).
$$\omega^{(n)} = 1/N : n = 1, ..., N$$
Using this initial set of weights, we find the initial CTBN structure $T_1$ (and its parameters $\bm\theta_{T_1}$) and set the current model $M$ to be $T_1$.
We then estimate the prediction error margin $\omega^{(n)} = 1 - P(\mathbf{y}^{(n)}|\mathbf{x}^{(n)},M)$ for each instance and renormalize such that $\sum_{n=1}^N \omega^{(n)} = 1$. We repeat our structure learning algorithm with $\{\omega^{(n)}\}$ and find another CTBN structure $T_2$.
After that, we set the current model to be the mixture of $T_1$ and $T_2$ and learn the ML-ME parameters $\{ \bm\Theta_G, \bm\Theta_T \}$ according to Algorithm \ref{alg:learn-ML-ME-parameters}.

We incrementally add trees to the mixture by repeating this process. To stop the process, we use internal validation approach. 
Specifically, the data used for learning are split to internal train and test sets. 
The structure of the trees and parameters are always learned on the internal train set. The quality of the current mixture is evaluated on the internal test set.
The mixture growth stops when the log-likelihood on the internal test set for the new mixture is worse than for the previous mixture.
The trees included in the previous mixture are then fixed, and the parameters of the mixture are relearned on the full training data.

\subsubsection{Complexity}
To learn a single CTBN structure, we need to compute edge weights for the complete graph $G$, which requires estimating $P(Y_i|\mathbf{X}, Y_j)$ for all $d^2$ pairs of classes. Finding the maximum branching in $G$ can be obtained in $O(d^2)$ using \citep{Tarjan:1977}. To learn $K$ CTBN structures for the mixture, we repeat these steps $K$ times. Therefore, the overall complexity is $O(Kd^2)$ times the complexity of learning logistic regression.

\subsection{PREDICTION}
\label{subsec:predict}

In order to make a prediction for a new instance $\mathbf{x}$, we want to find the MAP assignment of the class variables (see Equation (\ref{OPT-classification})). In general, this requires to evaluate all possible assignments of values to $d$ class variables, which is exponential in $d$. 

One important advantage of the CTBN model is that the MAP inference can be done more efficiently by avoiding blind enumeration of all possible assignments. More specifically, the MAP inference on a CTBN is linear in the number of classes ($O(d)$) when it is implemented using a variant of the max-sum algorithm on a tree structure \citep{batal:2013:CIKM}. 

However, our ML-ME model consists of multiple CTBNs and the MAP solution may, at the end, require enumeration of exponentially many class assignments. 
To address this problem, we rely on approximate MAP inference. The two commonly applied MAP approximation approaches in the literature are: convex programming relaxation via dual decomposition \citep{Sontag:thesis:2010}, and simulated annealing using a Markov chain \citep{Yuan:2004:UAI}. In this work, we use the latter approach. Briefly, we search the space of all assignments by defining a Markov chain that is induced by local changes to individual class labels. The annealed version of the exploration procedure \citep{Yuan:2004:UAI} is then used to speed up the search. We initialize our MAP algorithm using the following heuristic: first, we identify the MAP assignments for each CTBN in the mixture individually, and after that, we pick the best assignment from among these candidates. We have found this (efficient) heuristic to work very well and often results in the true MAP assignment.

\section{EXPERIMENTS}
\label{sec:experiments}

\subsection{DATA}

\begin{table}[b!]
\caption{ Datasets characteristics\\ \scriptsize{$N$: number of instances, $m$: number of features, $d$: number of classes, LC: label cardinality, DLS: distinct label set, DM: domain}}
\label{table:datasets}
\begin{center}
	{\fontsize{7.75}{9.5}\rm
	\begin{tabular}{ c  c  c  c  c  c  c }
		\textbf{DATASET} &  \textsc{$N$} &  \textsc{$m$} &  \textsc{$d$} &  \textbf{LC} & \textbf{DLS} & \textbf{DM}\\
		\hline
		Emotions & 593 & 72 & 6 & 1.87 & 27 & music\\
		Yeast & 2,417 & 103 & 14 & 4.24 & 198 & biology\\
		Scene & 2,407 & 294 & 6 & 1.07 & 15 & image\\
		Image & 2,000 & 135 & 5 & 1.24 & 20 & image\\
		Enron & 1,702 & 1,001 & 53 & 3.38 & 753 & text\\
		RCV1.sub1 & 6,000 & 8,394 & 10 & 1.31 & 69 & text\\
		RCV1.sub2 & 6,000 & 8,304 & 10 & 1.21 & 70 & text\\
		RCV1.sub3 & 6,000 & 8,328 & 10 & 1.22 & 74 & text\\
		RCV1.sub4 & 6,000 & 8,332 & 10 & 1.22 & 79 & text\\
		RCV1.sub5 & 6,000 & 8,367 & 10 & 1.31 & 76 & text\\
		\hline
	\end{tabular}
	}
\end{center}
\end{table}

We use ten publicly available MLC datasets obtained from different domains including music recognition, semantic image labeling, biology and text classification. Table \ref{table:datasets} summarizes the characteristics of the datasets. It shows the number of instances ($N$), the number of feature variables ($m$) and the number of class variables ($d$). In addition, it shows two statistics: 1) label cardinality (LC), which is the average number of labels per instance and 2) distinct label sets (DLS), which is the number of distinct class configurations that appear in the data. Note that for RCV1 datasets we have used the 10 most common labels.

\begin{table*}[t]
\caption{\footnotesize Performance of each method on the benchmark datasets in terms of exact match accuracy.\\
	\scriptsize{
	Marker $\ast$/$\circledast$ indicates whether ML-ME is statistically superior/inferior to the compared method (by paired t-test at $0.05$ significance level).
	}
}
\label{table:EMA-results}

\begin{center}
{\fontsize{7.75}{9.5}\rm
\begin{tabular}{|c||l|l|l|l|l|l|l|l|l|l|}
\hline
\textbf{EMA} & \multicolumn{1}{|c}{\textbf{BR}} & \multicolumn{1}{|c}{\textbf{CHF}} & \multicolumn{1}{|c}{\textbf{MLKNN}} & \multicolumn{1}{|c}{\textbf{IBLR}} & \multicolumn{1}{|c}{\textbf{CC}} & \multicolumn{1}{|c}{\textbf{ECC}} & \multicolumn{1}{|c}{\textbf{PCC}} & \multicolumn{1}{|c}{\textbf{MMOC}} & \multicolumn{1}{|c}{\textbf{CTBN}} & \multicolumn{1}{|c|}{\textbf{ML-ME}} \\ \hline
emotions           & 0.265 $\ast$                 & 0.300 $\ast$                  & 0.283 $\ast$                    & 0.335                     & 0.268 $\ast$                 & 0.288 $\ast$                  & 0.317                    & 0.332                     & 0.322                     & 0.353                       \\ \hline
image              & 0.280 $\ast$                 & 0.360 $\ast$                  & 0.346 $\ast$                    & 0.387 $\ast$                   & 0.426 $\ast$                 & 0.413 $\ast$                  & 0.449 $\ast$                  & 0.448 $\ast$                   & 0.414 $\ast$                   & 0.479                       \\ \hline
scene              & 0.541 $\ast$                 & 0.605 $\ast$                  & 0.629 $\ast$                    & 0.644 $\ast$                   & 0.632 $\ast$                 & 0.658 $\ast$                  & 0.666 $\ast$                  & 0.664 $\ast$                   & 0.625 $\ast$                   & 0.711                       \\ \hline
yeast              & 0.151 $\ast$                 & 0.163 $\ast$                  & 0.179 $\ast$                    & 0.204 $\ast$                   & 0.193 $\ast$                 & 0.204 $\ast$                  & 0.230                    & 0.219 $\ast$                   & 0.192 $\ast$                   & 0.244                       \\ \hline
enron              & 0.164 $\ast$                  & 0.170 $\ast$                   & 0.078 $\ast$                    & 0.163 $\ast$                   & 0.173 $\ast$                  & 0.180 $\ast$                   & -                        & -                         & 0.167 $\ast$                    & 0.196                       \\ \hline
rcv1.sub1 & 0.334 $\ast$                 & 0.357 $\ast$                  & 0.205 $\ast$                    & 0.279 $\ast$                   & 0.429 $\ast$                 & 0.410 $\ast$                  & 0.432 $\ast$                  & -                         & 0.441 $\ast$                   & 0.470                       \\ \hline
rcv1.sub2 & 0.439 $\ast$                 & 0.465 $\ast$                  & 0.288 $\ast$                    & 0.417 $\ast$                   & 0.516 $\ast$                 & 0.509 $\ast$                  & 0.523 $\ast$                  & -                         & 0.531                     & 0.536                       \\ \hline
rcv1.sub3 & 0.466 $\ast$                 & 0.486 $\ast$                  & 0.327 $\ast$                    & 0.446 $\ast$                   & 0.539 $\ast$                 & 0.539 $\ast$                  & 0.548 $\ast$                  & -                         & 0.560 $\ast$                   & 0.565                       \\ \hline
rcv1.sub4 & 0.510 $\ast$                 & 0.531 $\ast$                  & 0.354 $\ast$                    & 0.491 $\ast$                   & 0.579 $\ast$                 & 0.569 $\ast$                  & 0.588                    & -                         & 0.592                     & 0.592                       \\ \hline
rcv1.sub5 & 0.439 $\ast$                 & 0.456 $\ast$                  & 0.276 $\ast$                    & 0.411 $\ast$                   & 0.497 $\ast$                 & 0.494 $\ast$                  & 0.519 $\ast$                  & -                         & 0.539                     & 0.538                       \\ \hline
\end{tabular}
}
\end{center}
\end{table*}

\begin{table*}[t]
\caption{\footnotesize Performance of each method on the benchmark datasets in terms of conditional log-likelihood loss.\\
	\scriptsize{
	Marker $\ast$/$\circledast$ indicates whether ML-ME is statistically superior/inferior to the compared method (by paired t-test at $0.05$ significance level).
	}
}
\label{table:CLL-results}

\begin{center}
{\fontsize{7.75}{9.5}\rm
\begin{tabular}{|c||l|l|l|l|l|l|l|l|}
\hline
\textbf{CLL-Loss}      & \multicolumn{1}{|c}{\textbf{BR}} & \multicolumn{1}{|c}{\textbf{CHF}} & \multicolumn{1}{|c}{\textbf{MLKNN}} & \multicolumn{1}{|c}{\textbf{IBLR}} & \multicolumn{1}{|c}{\textbf{CC}} & \multicolumn{1}{|c}{\textbf{PCC}} & \multicolumn{1}{|c}{\textbf{CTBN}} & \multicolumn{1}{|c|}{\textbf{ML-ME}} \\ \hline
emotions           & 153.5 $\ast$                 & 147.5 $\ast$                  & 151.7 $\ast$                    & 143.0 $\ast$                   & 169.6 $\ast$                 & 134.9                    & 147.4 $\ast$                       & 131.3                        \\ \hline
image              & 432.5 $\ast$                 & 415.9 $\ast$                  & 425.3 $\ast$                    & 395.6 $\ast$                   & 480.3 $\ast$                 & 354.7                    & 388.4 $\ast$                       & 339.2                        \\ \hline
scene              & 344.7 $\ast$                 & 318.4 $\ast$                  & 310.9 $\ast$                    & 283.9 $\ast$                   & 395.0 $\ast$                 & 258.9 $\ast$                  & 306.3 $\ast$                       & 234.6                        \\ \hline
yeast              & 1500.3 $\ast$                & 1491.7 $\ast$                 & 1464.9 $\ast$                   & 1434.2 $\ast$                  & 2303.8 $\ast$                & 932.1 $\circledast$                & 1097.0 $\ast$                      & 989.5                        \\ \hline
enron              & 1290.0 $\ast$                     & 1272.5 $\ast$                 & 1301.2 $\ast$                   & 1287.4 $\ast$                  & 1293.5 $\ast$                & -                        & 1437.9 $\ast$                      & 1213.4                       \\ \hline
rcv1.sub1          & 1443.8 $\ast$                & 2144.2 $\ast$                 & 1873.7 $\ast$                   & 1379.5 $\ast$                  & 1701.3 $\ast$                & 1034.3 $\ast$                 & 962.7                         & 980.9                        \\ \hline
rcv1.sub2          & 1207.4 $\ast$                & 2223.6 $\ast$                 & 1687.8 $\ast$                   & 1172.6 $\ast$                  & 1398.8 $\ast$                & 923.0 $\ast$                  & 893.5 $\ast$                       & 858.1                        \\ \hline
rcv1.sub3          & 1207.4 $\ast$                & 2156.0 $\ast$                 & 1674.6 $\ast$                   & 1168.2 $\ast$                  & 1500.5 $\ast$                & 896.7 $\ast$                  & 939.7 $\ast$                       & 840.1                        \\ \hline
rcv1.sub4          & 1072.9 $\ast$                & 1759.9 $\ast$                 & 1532.9 $\ast$                   & 1034.8 $\ast$                  & 1282.1 $\ast$                & 823.0 $\ast$                  & 790.7 $\ast$                       & 770.7                        \\ \hline
rcv1.sub5	         & 1267.0 $\ast$                & 2283.6 $\ast$                 & 1795.5 $\ast$                   & 1234.7 $\ast$                  & 1422.0 $\ast$                & 1009.0 $\ast$                 & 924.0 $\ast$                       & 898.4                        \\ \hline
\end{tabular}
}
\end{center}
\end{table*}

\begin{table*}[t]
\caption{\footnotesize Performance of each method on the benchmark datasets in terms of micro F1.\\
	\scriptsize{
	Marker $\ast$/$\circledast$ indicates whether ML-ME is statistically superior/inferior to the compared method (by paired t-test at $0.05$ significance level).
	}
}
\label{table:microF1-results}

\begin{center}
{\fontsize{7.75}{9.5}\rm
\begin{tabular}{|c||l|l|l|l|l|l|l|l|l|l|}
\hline
\textbf{MICRO F1} & \multicolumn{1}{|c}{\textbf{BR}} & \multicolumn{1}{|c}{\textbf{CHF}} & \multicolumn{1}{|c}{\textbf{MLKNN}} & \multicolumn{1}{|c}{\textbf{IBLR}} & \multicolumn{1}{|c}{\textbf{CC}} & \multicolumn{1}{|c}{\textbf{ECC}} & \multicolumn{1}{|c}{\textbf{PCC}} & \multicolumn{1}{|c}{\textbf{MMOC}} & \multicolumn{1}{|c}{\textbf{CTBN}} & \multicolumn{1}{|c|}{\textbf{ML-ME}} \\ \hline
emotions  & 0.645 $\ast$                 & 0.672                    & 0.656 $\ast$                    & 0.692                     & 0.621 $\ast$                 & 0.652 $\ast$                  & 0.664                    & 0.687                   & 0.678                     & 0.694                       \\ \hline
image     & 0.479 $\ast$                 & 0.541 $\ast$                  & 0.504 $\ast$                    & 0.573                     & 0.550 $\ast$                 & 0.563 $\ast$                  & 0.565 $\ast$                  & 0.572 $\ast$                 & 0.561 $\ast$                   & 0.597                       \\ \hline
scene     & 0.696 $\ast$                 & 0.722 $\ast$                  & 0.736 $\ast$                    & 0.758                     & 0.697 $\ast$                 & 0.724 $\ast$                  & 0.722 $\ast$                  & 0.711 $\ast$                 & 0.717 $\ast$                   & 0.762                       \\ \hline
yeast     & 0.635                   & 0.637                    & 0.646                      & 0.661 $\circledast$                 & 0.628                   & 0.631                    & 0.645                    & 0.651                   & 0.631                     & 0.640                       \\ \hline
enron     & 0.551  $\ast$             & 0.569                    & 0.450 $\ast$                    & 0.566                     & 0.577                  & 0.583 $\circledast$                & -                        & -                       & 0.552                     & 0.565                       \\ \hline
rcv1.sub1 & 0.503 $\ast$                 & 0.516 $\ast$                  & 0.257 $\ast$                    & 0.459 $\ast$                   & 0.511 $\ast$                 & 0.525                    & 0.510 $\ast$                  & -                       & 0.512 $\ast$                   & 0.533                       \\ \hline
rcv1.sub2 & 0.568 $\ast$                 & 0.584 $\ast$                  & 0.317 $\ast$                    & 0.546 $\ast$                   & 0.586                   & 0.589                    & 0.588                    & -                       & 0.591                     & 0.590                       \\ \hline
rcv1.sub3 & 0.576 $\ast$                 & 0.592 $\ast$                  & 0.364 $\ast$                    & 0.564 $\ast$                   & 0.594                   & 0.610 $\circledast$                & 0.594                    & -                       & 0.596 $\ast$                   & 0.602                       \\ \hline
rcv1.sub4 & 0.622 $\ast$                 & 0.637                    & 0.404 $\ast$                    & 0.606 $\ast$                   & 0.640                   & 0.646 $\circledast$                & 0.644 $\circledast$                & -                       & 0.638                     & 0.634                       \\ \hline
rcv1.sub5 & 0.582 $\ast$                 & 0.597                    & 0.314 $\ast$                    & 0.566 $\ast$                   & 0.595                   & 0.603 $\circledast$                & 0.600 $\circledast$                & -                       & 0.598                     & 0.595                       \\ \hline
\end{tabular}
}
\end{center}
\end{table*}

\begin{table*}[t]
\caption{\footnotesize Performance of each method on the benchmark datasets in terms of macro F1.\\
	\scriptsize{
	Marker $\ast$/$\circledast$ indicates whether ML-ME is statistically superior/inferior to the compared method (by paired t-test at $0.05$ significance level).
	}
}
\label{table:macroF1-results}

\begin{center}
{\fontsize{7.75}{9.5}\rm
\begin{tabular}{|c||l|l|l|l|l|l|l|l|l|l|}
\hline
\textbf{MACRO F1} & \multicolumn{1}{|c}{\textbf{BR}} & \multicolumn{1}{|c}{\textbf{CHF}} & \multicolumn{1}{|c}{\textbf{MLKNN}} & \multicolumn{1}{|c}{\textbf{IBLR}} & \multicolumn{1}{|c}{\textbf{CC}} & \multicolumn{1}{|c}{\textbf{ECC}} & \multicolumn{1}{|c}{\textbf{PCC}} & \multicolumn{1}{|c}{\textbf{MMOC}} & \multicolumn{1}{|c}{\textbf{CTBN}} & \multicolumn{1}{|c|}{\textbf{ML-ME}} \\ \hline
emotions  & 0.632 $\ast$                 & 0.667                    & 0.656 $\ast$                    & 0.690                     & 0.620 $\ast$                 & 0.643 $\ast$                  & 0.659 $\ast$                  & 0.679                     & 0.670                     & 0.692                       \\ \hline
image     & 0.486 $\ast$                 & 0.546 $\ast$                  & 0.516 $\ast$                    & 0.581                     & 0.562 $\ast$                 & 0.571 $\ast$                  & 0.575 $\ast$                  & 0.578 $\ast$                   & 0.572 $\ast$                   & 0.605                       \\ \hline
scene     & 0.703 $\ast$                 & 0.730 $\ast$                  & 0.743 $\ast$                    & 0.765                     & 0.709 $\ast$                 & 0.740 $\ast$                  & 0.729 $\ast$                  & 0.721 $\ast$                   & 0.728 $\ast$                   & 0.772                       \\ \hline
yeast     & 0.457 $\ast$                 & 0.461                    & 0.478                      & 0.498 $\circledast$                 & 0.467                   & 0.477                    & 0.486                    & 0.473                     & 0.467                     & 0.472                       \\ \hline
enron     & 0.478                   & 0.479                    & 0.411 $\ast$                    & 0.475                     & 0.484                   & 0.482                    & -                        & -                         & 0.470                     & 0.483                       \\ \hline
rcv1.sub1 & 0.495 $\ast$                 & 0.511 $\ast$                  & 0.273 $\ast$                    & 0.463 $\ast$                   & 0.506 $\ast$                 & 0.516                    & 0.504 $\ast$                  & -                         & 0.507 $\ast$                   & 0.525                       \\ \hline
rcv1.sub2 & 0.503 $\ast$                 & 0.526 $\ast$                  & 0.264 $\ast$                    & 0.475 $\ast$                   & 0.531                   & 0.539                    & 0.531                    & -                         & 0.536                     & 0.536                       \\ \hline
rcv1.sub3 & 0.513 $\ast$                 & 0.536 $\ast$                  & 0.278 $\ast$                    & 0.497 $\ast$                   & 0.547                   & 0.558 $\circledast$                & 0.548                    & -                         & 0.543                     & 0.548                       \\ \hline
rcv1.sub4 & 0.499 $\ast$                 & 0.519                    & 0.269 $\ast$                    & 0.477 $\ast$                   & 0.534                   & 0.540 $\circledast$                & 0.534 $\circledast$                & -                         & 0.526                     & 0.522                       \\ \hline
rcv1.sub5 & 0.500 $\ast$                 & 0.526                    & 0.257 $\ast$                    & 0.487 $\ast$                   & 0.536 $\circledast$               & 0.538 $\circledast$                & 0.534 $\circledast$                & -                         & 0.536 $\circledast$                 & 0.528                       \\ \hline
\end{tabular}
}
\end{center}
\end{table*}

\subsection{METHODS}
We compare the performance of our proposed method, which we refer to as ML-ME, with the following MLC methods:

\vspace{-0.175in}
\begin{itemize}
	\itemsep-0.1em
	\item \emph{Binary Relevance (BR)} \citep{Boutell:2004:PR,Clare:2001:PKDD}
	\item \emph{Classification with Heterogeneous Features (CHF)} \citep{Godbole:2004:PAKDD}
	\item \emph{Multi-Label k-Nearest Neighbor (MLKNN)} \citep{Zhang:2007:PR}
	\item \emph{Instance-Based Learning by Logistic Regression (IBLR)} \citep{Cheng:2009:ML}
	\item \emph{Classifier Chains (CC)} \citep{Read:2009:ECML}
	\item \emph{Ensemble of Classifier Chains (ECC)} \citep{Read:2009:ECML}
	\item \emph{Probabilistic Classifier Chains (PCC)} \citep{Dembczynski:2010:ICML}
	\item \emph{Maximum Margin Output Coding (MMOC)} \citep{Zhang:2012:ICML}
	\item \emph{Single CTBN (CTBN)} \citep{batal:2013:CIKM}
\end{itemize}

For all methods, we use the same parameter settings as suggested in the papers that introduced them: 
For MLKNN and IBLR, we use Euclidean distance to measure similarity of instances and we set the number of nearest neighbors to 10 \citep{Zhang:2007:PR,Cheng:2009:ML}; 
for CC, we set the order of classes to $Y_1\!<\!Y_2,... <\!Y_d$ \citep{Read:2009:ECML};
for ECC, we use 10 CCs in the ensemble \cite{Read:2009:ECML};
and for MMOC, we set $\lambda$ (the decoding parameter) to 1 \citep{Zhang:2012:ICML}.
Also note that all of these methods except MMOC are considered as meta-learners because they can work with several base classifiers.
To eliminate additional effects that may bias the results, we use $L_2$-penalized logistic regression for all of these methods and choose their regularization parameters by cross validation. Lastly, for our ML-ME model, we set the number of mixture components to 5 and we use 150 iterations of simulated annealing for prediction.

\subsection{EVALUATION MEASURES}

Evaluating the performance of MLC methods is more challenging than that of traditional single-label classification methods. The most appropriate measure is the \textit{exact match accuracy} (EMA), which computes the percentage of instances whose predicted output vectors are exactly the same as their true class vectors (i.e., all classes are predicted correctly).
This measure is proper for MLC because it evaluates the success of the method in finding the mode of $P(\mathbf{X}|\mathbf{Y})$ (see Section \ref{sec:problem_definition}).
However, EMA could be too harsh, especially when the output dimensionality is high.
An alternative evaluation measure is the \textit{conditional log-likelihood loss} (CLL-loss), which computes the negative conditional log-likelihood of the test instances:
$$\mbox{\footnotesize \emph{CLL-loss}}=\sum_{n=1}^{N} -\log \left( P(\mathbf{y}^{(n)}|\mathbf{x}^{(n)}) \right)$$
CLL-loss evaluates how much probability mass is given to the true label vectors (the higher the probability, the smaller the loss). Note that CLL-loss is only defined for probabilistic methods.

\textit{Micro F1} and \textit{macro F1} have been used for evaluating MLC methods \citep{Tsoumakas:2009:PKDD}. Micro F1 aggregates the number of true positives, false positives and false negatives for all classes and then calculates the overall F1 score. On the other hand, macro F1 computes the F1 score for each class separately and then averages these scores.
Note that both measures are not quite appropriate for MLC because they do not account for the correlations between classes \citep{Dembczynski:2010:ICML}. However, we report them in our results as they have been used in other MLC literature.

\subsection{RESULTS}

Tables \ref{table:EMA-results}, \ref{table:CLL-results}, \ref{table:microF1-results} and \ref{table:macroF1-results} 
show the performance of all methods in terms of EMA, CLL-loss, micro F1 and macro F1, respectively. All results are obtained using \textit{ten-fold cross valiation}. To evaluate the statistical significance in the performance measure differences, we use paired t-test at 0.05 significance level. We use markers $\ast$/$\circledast$ to indicate whether ML-ME is statistically superior/inferior to the compared method.

Note that we report the results of MMOC only on four datasets (emotions, image, scene and yeast) because it did not finish on the rest of the datasets (MMOC did not finish one round of the learning within 24 hours). Also, PCC did not finish on the enron dataset be causes it has to evaluate all $2^{53}$ possible class assignments, which is clearly infeasible. Lastly, we do not report CLL-loss for for MMOC and ECC because they do not compute the probabilistic score for  a giving class assignment.

In terms of EMA (Table \ref{table:EMA-results}), ML-ME clearly outperforms the other methods on most datasets. For example, ML-ME is significantly better than BR, CHF, MLKNN, CC and ECC on all ten datasets, significantly better than IBLR on nine datasets, significantly better than PCC on six datasets and significantly better than MMOC on three datasets.
In addition, ML-ME shows significant improvement over a single CTBN on six datasets, which demonstrates the ability of the mixture to compensate for the tree structure restriction of a single CTBN.

In terms of CLL-loss (Table \ref{table:CLL-results}), ML-ME again shows consistent improvement over the other methods.
This is because ML-ME is learned to optimize the log-likelihood of the data and its prediction is based on exact MAP inference.
Also note that ML-ME shows a consistent improvement over CTBN because combining multiple CTBNs allows us to account for different relations in the data and, hence, improves the generalization of the model.
%
%

Lastly, Table \ref{table:microF1-results} and \ref{table:macroF1-results} show that ML-ME is also very competitive in terms of micro and macro F1 scores, although optimizing them was not our main objective. 

In summary, the experimental results show that our ME-ML method with CTBN experts is able to outperform or match the existing state-of-the-art methods across a broad range of benchmark MLC datasets. We attribute this performance to the ability of CTBN mixtures to simultaneously compensate for the restricted dependences modeled by an individual CTBN, and for its ability to fit better the different regions of the input space with new expert models.

\section{CONCLUSION}
\label{sec:conclusion}

In this paper, we have proposed a new probabilistic approach for the multi-label classification problem. 
Our approach models different input-output relations using conditional tree-structured Bayesian networks, while the mixtures-of-experts architecture aims to compensate for the tree-structured restrictions and as a result achieves a more accurate model.
We have formulated and developed the algorithms for learning the model from data and for performing multi-label predictions on future data instances.
Our experiments on a broad range of datasets showed that our approach outperforms several state-of-the-art methods and produces more reliable probabilistic estimates.

%

\subsubsection*{Referencess}
\vspace{-0.285in}
\bibliographystyle{plainnat}
\begin{small}

\end{small}
\end{document}